\documentclass[letterpaper, 10 pt, journal, twoside]{IEEEtran} 
\usepackage{blindtext, graphicx}
\usepackage{tabularx}
\usepackage[table,xcdraw]{xcolor}
\usepackage{multirow}
\usepackage{amsmath}
\usepackage{amssymb}
\usepackage{caption}
\usepackage{authblk}
\usepackage{multicol}
\usepackage[tight,footnotesize]{subfigure}
\usepackage[sort]{cite}
\usepackage[none]{hyphenat} 
\usepackage[utf8]{inputenc}
\DeclareGraphicsExtensions{.png,.pdf}

\usepackage{float}
\usepackage{placeins}

\usepackage{xcolor}  
\usepackage{soul}   
\usepackage[inline]{enumitem}
\usepackage{hyperref}
\usepackage{algpseudocode}
\usepackage{algorithm}
\algnewcommand{\algorithmicand}{\textbf{ and }}
\algnewcommand{\algorithmicor}{\textbf{ or }}
\algnewcommand{\OR}{\algorithmicor}
\algnewcommand{\AND}{\algorithmicand}

\soulregister\cite7
\soulregister\ref7
\soulregister\pageref7

\renewcommand{\vec}[1]{{\boldsymbol{#1}}}

\author{Riccardo Polvara$^{*}$$^{1}$, Francesco Del Duchetto$^{*}$$^{1}$$^{2}$,  Gerhard Neumann$^{3}$ and Marc Hanheide$^{1}$%
\thanks{Manuscript received: February, 24, 2021; Revised May, 24, 2021; Accepted June, 21, 2021.}
\thanks{This paper was recommended for publication by Editor Youngjin Choi upon evaluation of the Associate Editor and Reviewers' comments.
This work was supported by EPSRC under grant agreement EP/R02572X/1 (National Center for Nuclear Robotics), the EU's H2020 research and innovation program under grant agreement No 871704 (BACCHUS) and partially funded by Berry Garden Growers.} 
\thanks{$^{*}$Both first and second author contributed equally and should be considered co-first authors.}
\thanks{$^{1}$Authors are with Lincoln Center for Autonomous Systems (L-CAS), School of Computer Science, University of Lincoln, United Kingdom.{\tt \footnotesize \{rpolvara, fdelduchetto, mhanheide\}@lincoln.ac.uk} }
\thanks{$^{2}$Saga Robotics, Lincoln, UK}
\thanks{$^{3}$Karlsruhe Institute of Technology, Karlsruhe, Germany. \tt \footnotesize {gerhard.neumann@kit.edu}}
\thanks{Digital Object Identifier (DOI): see top of this page.}
}

\title{Navigate-and-Seek: a Robotics Framework for People Localization in Agricultural Environments}

\begin{document}
\maketitle
\begin{abstract}
The agricultural domain offers a working environment where many human laborers are nowadays employed to maintain or harvest crops, with huge potential for productivity gains through the introduction of robotic automation. Detecting and localizing humans reliably and accurately in such an environment, however, is a prerequisite to many services offered by fleets of mobile robots collaborating with human workers.
Consequently, in this paper, we expand on the concept of a topological particle filter (TPF) to accurately and individually localize and track workers in a farm environment, integrating information from heterogeneous sensors and combining local active sensing (exploiting a robot's onboard sensing employing a Next-Best-Sense planning approach) and global localization (using affordable IoT GNSS devices).
We validate the proposed approach in topologies created for the deployment of robotics fleets to support fruit pickers in a real farm environment. 
By combining multi-sensor observations on the topological level complemented by active perception through the NBS approach, we show that we can improve the accuracy of picker localization in comparison to prior work.
\end{abstract}


%
\IEEEpeerreviewmaketitle

\section{Introduction}
\label{introduction}
\IEEEPARstart{B}{eing} able to reliably and precisely maintain a representation of the location of humans is considered a prerequisite to allow humans and robots to work collaboratively in a shared environment. 

\begin{figure}
  \centering
  {\includegraphics[width=\columnwidth]{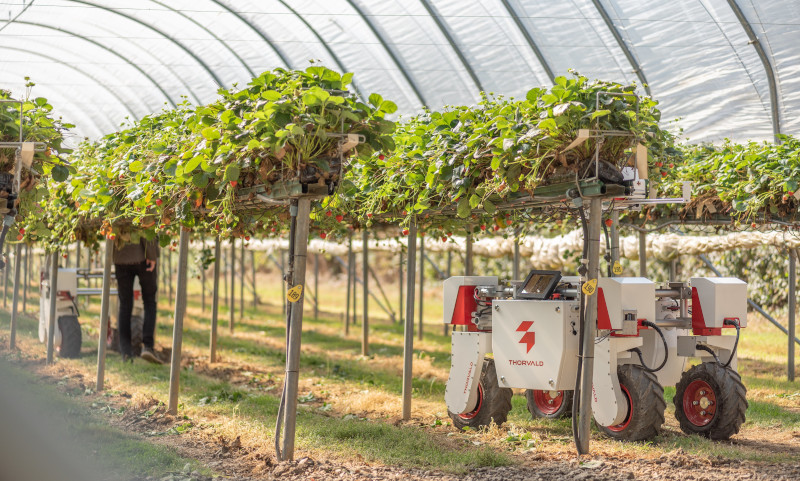}}
  \caption{The Thorvald robot is used for transportation in strawberry production farms.}
  \label{fig:thorvald}
\end{figure}

In this paper, we propose a framework that integrates passive (opportunistic) and active (explorative) sensing using a variety of sensors (GNSS, LIDAR, and RFID) into a Bayesian tracking framework operating on a topological representation of large-scale outdoor environments. 
The work is motivated by the sparsely connected topology typically found in many farm environments, where the movement of humans (and robots) is largely constrained by the physical structures, i.e., rows of plant-supporting tables in a so-called poly-tunnel as seen in Fig.~\ref{fig:thorvald}.
Such sparsely connected topological maps allow to constrain the movements of hypotheses along the defined topological edges (see Fig. 2 (bottom)), in order to provide a more accurate overall localization estimate, compensating for substantial noise of sensors, as first proposed by Khan \emph{et al}~\cite{khan2020incorporating}. 
It shall be noted that the benefits of such an approach\footnote{The implementation of our TPF framework is released open-source: \href{https://github.com/pulver22/nbs\_ros/tree/topologicalMap}{https://github.com/pulver22/nbs\_ros/tree/topologicalMap}} are not limited to agricultural robotics, but can be utilized in any tracking application that features a sparsely connected topological map.

The initial work~\cite{khan2020incorporating} solved the problem employing a ``Topological Particle Filter'' (TPF) and focused entirely on exploiting such sparsely linked topological graphs to compensate for the noise and inaccuracies of deployed cheap GNSS IoT sensors or mobile phones carried by human workers, to allow them to summon a robot to their locations as part of a robotic in-field logistics solution. 
This paper takes this idea and the TPF approach further, making the following tangible contributions:
\begin{enumerate*}[label=(\roman*)]
    \item Generalization of the TPF framework to support a larger variety of information sources to be fused, i.e., complementing noise GNSS information with detections of humans from a robot's onboard sensing such as LIDAR-based and RFID identification;
    \item complementing the opportunistic and passive sensing approach proposed in \cite{khan2020incorporating} with an intelligent decision-making active sensing approach known as Next-Best-Sense (NBS)~\cite{polvara2020next} to utilize the onboard sensors of the robotic platforms to further reduce the uncertainty of humans' location in the field, i.e., to actively move robots around the environment to systematically reduce sensing uncertainty;
    \item a comprehensive evaluation of the proposed integrated approach in a digital twin of the real-world logistics scenario, in which a fleet of robots is supporting human fruit pickers in soft-fruit production, by allowing them to focus on the picking of fruits themselves, while the robotic fleet automates the transportation of picked fruits from the human picker to a local storage facility.
\end{enumerate*}

The remainder of the article is organized as follows: 
in Sec.~\ref{state} we survey the related literature while 
in Sec.~\ref{sec:problem-form} we define the problem statement. In Sec.~\ref{sec:tpf} and Sec.~\ref{sec:nbs} we present the updated formulation of our topological particle filter and Next-Best-Sense. In Sec.~\ref{experiments} we report the results of experiments performed.
Finally, in Sec.~\ref{conclusion} we summarize our approach, highlighting the main concepts, its advantages and limitations, and future works.

\section{Related Work}\label{state}

This work focuses on the localization and tracking of fruit pickers in polytunnel environments and on the planning of an optimal navigation path that optimizes multiple criteria, such as localization accuracy and battery usage.    

\paragraph{People Localization and Tracking}\label{subsec:localization}

The problem of people localization and tracking has been studied in the literature from various angles in different fields.
Most of the time, the problem is characterized by the trade-off existing between a very high localization accuracy and very expensive sensors used to obtain such performance. Examples of expensive devices for accurate tracking are 3D dense LIDARs~\cite{Weng20203DMT}, high-resolution images, or "Real-Time-Kinematic" Global Navigation Satellite System (RTK-GNSS). One of the most cost-effective sensing approaches relies on color video streams~\cite{park2013face} which unfortunately lacks in providing reliable spatial information. Such limitation has been compensated by the adoption of RGBD cameras, which provide depth information in addition to the traditional color stream. For example, in \cite{gritti2014kinect} a small robot can track people in a cluttered indoor environment solely based on the images acquired by a Microsoft Kinect. 
Recently, many vision-based approaches have yielded promising results exploiting the power of Deep Learning for detecting and tracking moving objects \cite{Osep2018TrackTD, Sharma2018BeyondPL, Xu2020SegmentAP}.
Devices such as lasers and cameras cannot often identify a person within a group, even though some studies \cite{bellotto2007multisensor, bellotto2018human} presented a Bayesian approach for addressing this challenge. On the other hand, RFID antennas, given their ability to recognize uniquely identified tags, can be deployed to distinguish people in a crowd. In \cite{li2017online}, the authors combined the RFID readings with the human skeleton extracted by a video stream acquired with an RGBD camera. The main limitation of this work is that it can only extract up to six skeletons per frame. To solve this problem, \cite{yan2017online} adopts a multi-tracking learning system based on 3D LIDARs, unfortunately introducing a consistent number of false-positives.

Most of the methods described require the robot to be in close vicinity with the tracked agent to guarantee satisfying precision. This may work in a scenario of a limited size such as a supermarket or a hospital but fails in outdoor environments. In the agriculture domain,
\cite{Vougioukas2016OrchardWL} proposes an approach to localize workers in a farm using an ultra-wideband (UWB) radio-based system requiring, however, expensive specialized equipment.
Here, the adoption of a GNSS-based system still proves to be the most effective one. In~\cite{khan2020incorporating}, a particle filter is proposed for localizing humans over a topological map. Because it relies only on the GNSS signal, the performance is limited by the bias affecting the GNSS signal itself. 
Another limitation of this approach is that particles are only allowed to move along the topological edges, making the filter unable to recover from a wrong initialization. The method also assumes a fixed velocity model of the humans' dynamics which does not scale well when people move with different speed and directions.
In the current work, we build from~\cite{khan2020incorporating} attempting to solve the above issues by allowing to integrate multiple sensor modalities (rather than only GNSS),
estimating the velocity of targets alongside their position, and implementing a mechanism to allow particles to recover from a wrong initialization.

\paragraph{Active Robotic Sensing}\label{subsec:navigation}
The majority of exploration strategies for initially unknown environments greedily take decisions and are often called Next-Best-View (NBV) algorithms. In such systems, the next location for the robot is chosen to be on the boundary between the already explored free space and the unknown area. This decision is usually performed using a utility function. Popular functions are the \textit{traveling cost}~\cite{yamauchi}, according to which the next best observation location is the nearest one, and the \textit{information gain}~\cite{gonzales, stachniss}, defined as the expected amount of new information the robot can acquire from the candidate location. In~\cite{8737705}, the two are instead combined with semantic information. More recently, frontier-based exploration has been combined with sampling methods to prevent the robot gets stuck~\cite{8633925}. The limitation of the aforementioned methods is that they rely on \textit{ad-hoc} aggregation methods. To provide more theoretical foundation, Multi-Criteria Decision Making has been introduced in robotics from the information theory field \cite{mcdm}.
Only more recently we see the adoption of data-driven approaches for the exploration task. For example, long-term memory is used in \cite{zhang2017neural} to learn a global map starting from raw sensory inputs and reactive actions. Imitation~\cite{chen2019learning} and reinforcement learning~\cite{8606991} are also other new techniques that seem promising to address the task of interest. They also make a decision solely based on the partial information obtained by the robot. However, their use is still limited due to the large number of samples required for training the model.

\section{Problem Formulation}\label{sec:problem-form}
We specifically study the problem of accurately tracking humans using GPS, LIDAR, and RFID sensors. 
In our agricultural setting, humans move along polytunnels to harvest fruits and performing husbandry tasks. By sharing the environment with the humans, robots need to be aware of humans' positions for assisting their operations, similar to what can be seen in Fig. 2 (top). 
For example, in a fruit picking scenario in which the robot serves as a carrier from the human picker to the farm storage, the robot needs to know the human position to enter the correct polytunnel and reach them. A failure in identifying the correct lane means that the robot must navigate until the end before entering the correct one, or that the human has to reach the robot. All of which results in large delays.
The limited range of the sensors onboard requires the robot to get close to the humans to detect them accurately. Therefore, a navigation strategy is needed to improve the quality of the sensing operations while balancing other criteria like the traveled distance and the battery consumption.
\begin{figure}
  \centering
  \begin{minipage}[b]{\columnwidth}
      \subfigure{\label{fig:gazebo}\includegraphics[width=\textwidth]{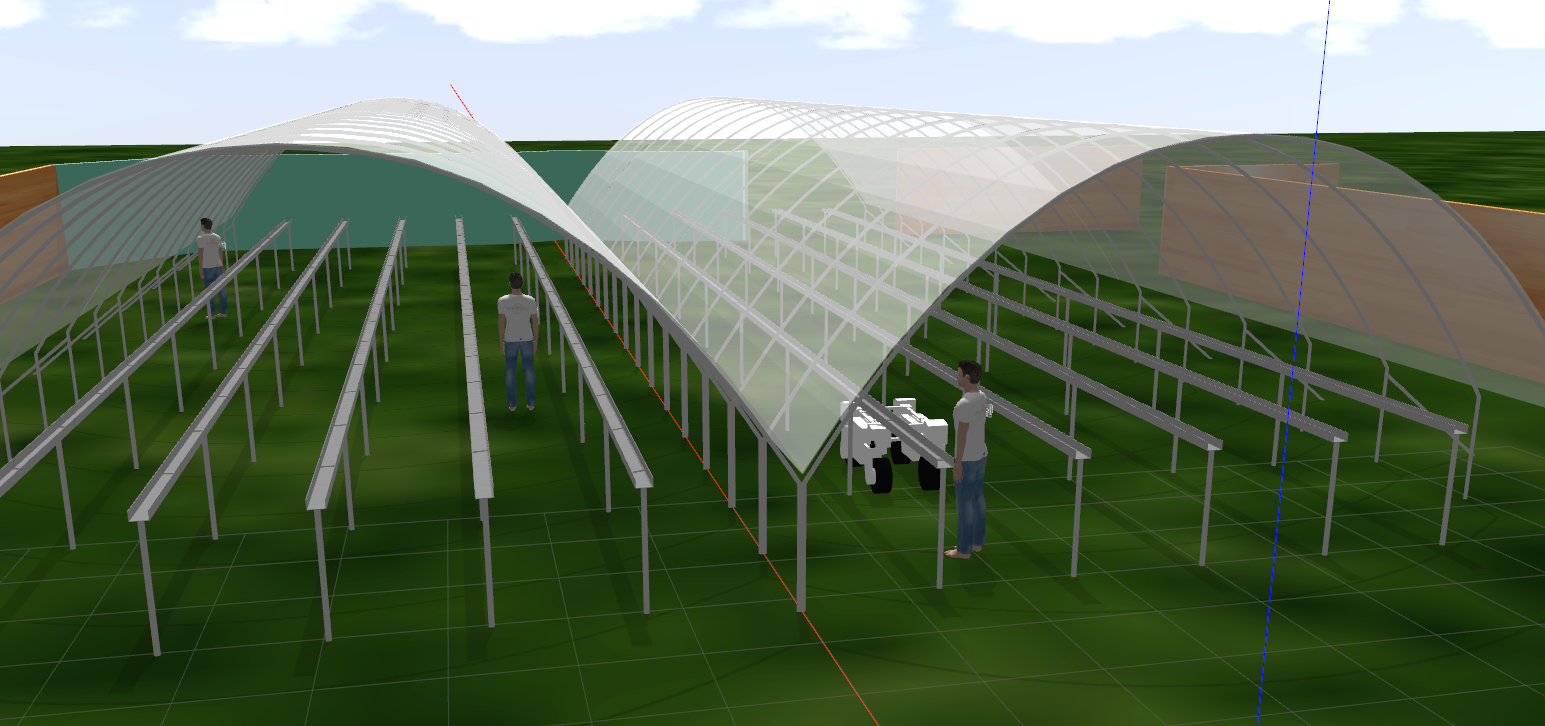}}
  \end{minipage}
  
  \begin{minipage}[b]{\columnwidth} 
    \subfigure{\label{fig:rviz}\includegraphics[width=\textwidth]{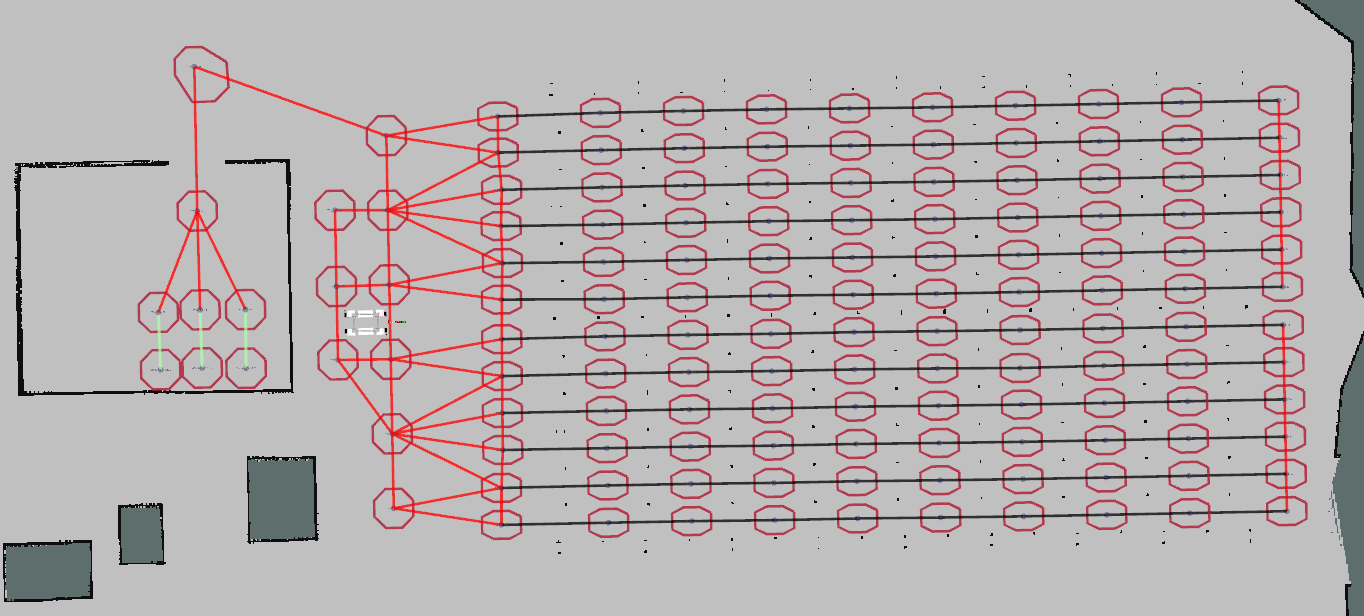}}
  \end{minipage}
  \caption{The Riseholme campus. (Top) A view of the Gazebo simulation built in Gazebo. (Bottom) The topological map of the environment used in our study.}
  \label{fig:simulated_scenario}
\end{figure}
Sensing operations are defined as \emph{identifying observations} if they uniquely identify a specific instance of the person tracked. 
Observations from GPS sensors are \emph{identifying} because, to use the technology, people have to carry with them an IoT GNSS device or a smartphone which can both be uniquely identified by their MAC addresses. Similarly, the RFID antenna on the robot can sense RFID tags in the environment which broadcast a unique ID. LIDAR sensors, instead, cannot identify the people detected.

The environment structure is represented as a graph, which we also refer to as \textit{topological map}. A topological map is a discrete representation that can be viewed as a tuple $T \longrightarrow \langle N; E \rangle$, where $N \in \mathbb{R}^2$ is a set of discrete physical locations in the Euclidean space, i.e. $n = \langle x, y \rangle, n \in N$, called \textit{topological nodes}.
The set $E \subseteq N \times N$ represents the set of possible edges connecting the topological nodes, where the element at $j$-th row and $k$-th column of $E$ is defined as $e_{jk} = 1$, if $n_j$ connects to $n_k$, $0$ otherwise. 
In this work, we approximate the humans' position on the field with the node closest to them. We assume that humans move by traversing the edges to go from one node to another. Nodes and edges are positioned in the environment to represent the locations over which humans and robots can navigate, therefore $N$ and $E$ act as inputs to the TPF to constrain the prediction of the particles' movements. 
The robot pose $c$ is defined by the topological node $n$ where it is located and by its orientation $\theta$. The robot can move from node $n$ to any of the nodes $n'$ connected to it, and perform a \emph{sensing operation} to locate the presence of any fruit picker in its surrounding. The problem of planning a path consists in finding the optimal sequence of sensing operations $\langle ((n_{1},\theta_{1})), ((n_{2},\theta_{2})), \ldots, ((n_{n},\theta_{n}))\rangle$ to be performed in order to localize all the fruit pickers in the environment.

\section{The Topological Particle Filter}\label{sec:tpf}


The Topological Particle Filter (TPF), introduced in  \cite{khan2020incorporating} and extended in this work, is a method for tracking the position of targets on a map that exploits the structure of the environment -- the topological map -- to find the closest node to the targets. 
The distribution of particles $P$ over the topology is used to approximate the probability distribution of the targets on the map.
At timestep $t$, each particle $p_t^i \in P_t$, for $i = 1,2,...,|P_t|$, in our updated TPF formulation is characterized by the state
\begin{equation*}
    p_{t}^i  = <q_t^i, \vec{v}_t^i, \tau_t^i, T, V> ,
\end{equation*} 
where $q_t^i \in N$ is the node the particle lies in, $\vec{v}_t^i$ is the velocity vector of the particle, 
$\tau_t^i$ is the amount of time particle $p_t^i$ has been in $q_t^i$, $T$ is the topology and $V$ is a fixed window size used to estimate the particle's velocity.
The definition of the particle uses the Markov assumption, i.e., a particle state at time $t+1$ depends uniquely on its state at time $t$.
Differently from~\cite{khan2020incorporating}, we want our filter to work with different sensors $S$. 
Therefore, without loss of generality, each observation from sensor $s \in S$ is passed to the filter in the form of a likelihood distribution $\mathcal{L}(N)$ of the target over the topological nodes $N$. Every observation is paired with a variable $\textrm{id}_s \in \{\top,\bot\}$ which indicates whether the observation coming from $s$ identifies the target or not (see Section \ref{sec:problem-form}).
Upon receiving an observation, the TPF updates its belief distribution of the target through the sequential steps of \emph{Prediction}, \emph{Weighting} and \emph{Resampling}, generating the estimate $n^*$ for the target position in the topological map.

\begin{algorithm} \caption{TPF update}\label{alg:tpf-algorithm}
\begin{algorithmic}[1]
\Require $\mathcal{L}(N)$, the sensor observation; $\textrm{id}_s$, the identifying variable; $P_t$, the current set of particles state; $\textrm{pr}_j$, the probability of jumping to non-connected nodes; $\epsilon_{JSD}$ and $\epsilon_{H}$, thresholds for monitoring the distribution of particles.
\If {$t = 0$}
    \State $P_t \gets Initialization(\mathcal{L}(N), \textrm{id}_s)$
\EndIf
\State $\acute{P}_{t{+1}} \gets Prediction(P_t)$
\State $d \gets \textrm{JSD}(\acute{P}_{t{+1}}, \mathcal{L}(N))$ \Comment{Jensen-Shannon Distance}
\If{$d > \epsilon_{JSD}$ \AND $\textrm{id}_s = \top$}
    \State $\acute{P}_{t{+1}} \gets Initialization(\mathcal{L}(N), \textrm{id}_s)$
    \State $\textrm{pr}_j \gets 1\mathrm{e}{-3}$ \Comment{Jumping to non-connected nodes}
\EndIf
\State $W, n_{t+1}^*\gets Weighting(\acute{P}_{t{+1}}, \mathcal{L}(N))$
\State $P_{t{+1}} \gets Resampling(\acute{P}_{t{+1}}, W, \textrm{pr}_j)$
\State $h_p \gets \textrm{H}(P_{t{+1}})$ \Comment{Entropy}
\If {$h_p < \epsilon_{H}$}
    \State $\textrm{pr}_j \gets 0$ \Comment{Only jump to connected nodes}
\EndIf
\State \Return $P_{t+1}$, $n_{t+1}^*$, $\textrm{pr}_j$
\end{algorithmic}
\end{algorithm}
Typical problems of naively performing the update steps of the TPF with noisy observations are that of wrong initialization~\cite{khan2020incorporating} and false-positive detections, biasing the distribution of the particles away from the real target position. To overcome these problems we have implemented two mechanisms. The first ensures, with a small probability, that the particles are allowed to jump to nodes not connected by an edge only when the entropy of the belief is above a certain confidence threshold. The second re-initializes the belief with the current \emph{identifying} observation in case the divergence between the observation (representing a true positive) and the belief is too large.
The divergence is computed as the Jensen-Shannon Distance
between the nodes distribution $Q(n) = \sum_{\forall i : q^i = n} 1 / \sum{P}$ of the particles and the distribution $L(n) = \mathcal{L}(n)/ \sum{\mathcal{L}(N)}$ of the observation
\begin{equation*}
    \textrm{JSD} = \sqrt{\frac{D_{KL}(Q\|A) + D_{KL}(A\|L)}{2}},
\end{equation*}
where $A$ is the point-wise mean between $Q$ and $L$, and $D_{KL}$ is the Kullback-Leibler divergence. We preferred the JSD over other divergences (e.g., the $D_{KL}$) because it always assumes a finite value within the interval $[0, 1]$.
Similarly, the entropy of the nodes distribution of $P$ is computed as
\begin{equation*}
    \textrm{H} = - \sum_{i=0}^{\mid N\mid}{Q(n_i)\log(Q(n_i))}.
\end{equation*}


The overall update function of the TPF is outlined in Algorithm \ref{alg:tpf-algorithm}, while the following paragraphs detail the implementation of the steps in the TPF update.

\paragraph{Initialization}
The state components of the particles are initialized with
\begin{equation}
    \begin{split}
        q_{t=0}& \sim \mathcal{L}(q_{t=0})/ \sum{\mathcal{L}(N)},\\
        v_{t=0}^x, v_{t=0}^y& \sim \mathcal{N}(\mu_{init}, \sigma^2_{init}),\\
        \tau_{t=0}& \sim \mathcal{U}(\tau^{-}_{init}, \tau^{+}_{init}).
    \end{split}
\end{equation}
In the case that $\textrm{id}_s = \bot$,
$\mathcal{L}(N)$ 
is artificially set to a uniform distribution over the entire map to avoid biasing the initial estimate with a possible false-positive.


\paragraph{Prediction}
Given the previous particle $p_{t}$, the next state update function predicts the state of the current particle $\acute{p}_{t{+1}}$.

The node state of the particle is predicted with a two-stage process. First, we decide whether the particle should jump from the current node to one of the nodes in $K = \{q_k \in N : e_{tk}=1\}$ connected to it, based on how long the particle has stayed in $q_{t}$ and the position of the nodes in $K$. This probability is defined as
\begin{multline}
    \textrm{Pr}(\acute{q}_{t+1} \neq q_t \mid q_t, \vec{v}_t, \tau_t, K) =\\ \frac{1}{\sum_k b_k}  \sum_{k \in K}{b_k\cdot (1 - \exp^{(\lambda(q_t,q_k,\vec{v}_t)\cdot \tau_t)})},
\end{multline}
where $b_k$ is a weighting factor equal to the projection of the speed of the particle $\vec{v}_t$ on the edge $e_{tk}$, i.e., $b_k = \max(0, \textit{proj}^{t,k}_{\vec{v}_t})$, with $\textit{proj}$ being the projection operation. This factor gives more probability to nodes that are on edges aligned with $\vec{v}_t$ and assigns null probability to those in the opposite direction. 
In the second stage, if a particle has been selected to jump, we predict which of the nodes connected to $q_{t}$ it should jump to. The probability is computed as a softmax  over the nodes in $K$
\begin{equation}
\textrm{Pr}(\acute{q}_{t+1}\mid q_t,\vec{v}_t, \tau_t, K) = \frac{\exp^{(\lambda(q_t,\acute{q}_{t+1},\vec{v}_t)\cdot \tau_t)}}{\sum_k \exp^{(\lambda(q_t,q_k,\vec{v}_t)\cdot \tau_t)}}.
\end{equation}
The function $\lambda$ depends on the speed of the target and the distance between the nodes in the topology. It is designed to have $50\%$ un-normalized probability of jumping to the next node $\acute{q}_{t+1}$ when the target is at equal distance between the current node $q_t$ and $\acute{q}_{t+1}$, i.e., when the time of particle in node $q_t$ is 
\begin{equation}\label{time}
\tau_t = \frac{d(\acute{q}_{t+1}, q_t)}{2 \cdot \max(0, \textit{proj}^{t,t+1}_{\vec{v}})},
\end{equation}
we want
\begin{equation}\label{unn}
\exp^{(\lambda(q_t,\acute{q}_{t+1},\vec{v}_t)\cdot \tau_t)} = \frac{1}{2}.
\end{equation}
Therefore substituting (\ref{time}) in (\ref{unn}) we get that
\begin{equation}\label{eq:prob_jump}
\lambda(q_t, \acute{q}_{t+1},\vec{v}) = \frac{2 \cdot \log(\frac{1}{2}) \cdot \max(0, \textit{proj}^{t,t+1}_{\vec{v}})}{d(\acute{q}_{t+1}, q_t)}.
\end{equation}
This formulation of $\lambda$ ensures that the TPF is robust to changes in topologies and movement patterns of the target, without having to resort to manually adjusted parameters.

Now, having defined the probability function for the node state prediction of the particle we can generate the current predicted state $\acute{p}_{t+1}$ having the following components
\begin{equation}
    \begin{split}
   & \acute{q}_{t{+1}} \sim \textrm{Pr}(\acute{q}_{t{+1}}\mid q_{t}, \vec{v}_{t}, \tau_{t}, T),\\
   & \acute{\tau}_{t{+1}} = \begin{cases}
              \tau_{t} + (\textrm{ts}_{t+1} - \textrm{ts}_t) & \text{if $\acute{q}_{t{+1}} = q_{t}$},\\
              0 & \text{otherwise},
            \end{cases} \\
   & \acute{\vec{v}}_{t{+1}} = \vec{v}_{t} + \textrm{Pr}(\acute{q}_{t{+1}}\mid q_{t}, \vec{v}_{t}, \tau_{t}, T) \cdot \frac{\frac{d(\acute{q}_{t+1}, q_t)}{\acute{\tau}_{t{+1}} - \tau_{t}} - \vec{v}_{t}}{V},
    \end{split}
\end{equation}
with $d(\acute{q}_{t+1}, q_t)$ the Euclidean distance between nodes $\acute{q}_{t+1}$ and $q_{t}$, and $\textrm{ts}_t$ the time in seconds at step $t$.
\paragraph{Weighting}
Upon receiving an observation from a sensor, we weight each particle $\acute{p}_{t+1}$ according to how much its predicted state agrees with what it has been observed. The weight is the sum of two factors $w_q$ and $w_{\vec{v}}$ which represent the confidence for the node and speed components of the particle's state respectively, i.e.,
\begin{equation}
    \begin{split}
        w_q = & \gamma_q^s \cdot \mathcal{L}(\acute{q}_{t{+1}}),\\
         w_{\vec{v}} = & 
                \gamma_{\vec{v}}^s  \cdot  \frac{1}{4} \Big(g_{\mu=\|\vec{v}^s\|,\sigma=\frac{\|\vec{v}^s\|}{2}}(\|\acute{\vec{v}}_{t{+1}}\|) \\&\quad\quad\quad+ \frac{\cos{(\angle{(\acute{\vec{v}}_{t{+1}},\vec{v^s})})} + 1}{2}\Big),
    \end{split}
\end{equation}
where $g_{\mu,\sigma}(x)$ is the Gaussian density with mean $\mu$ and variance $\sigma^2$, $\gamma_q^s$ and $\gamma_{\vec{v}}^s$ are weighting factors defined for each sensor $s$ providing observations, and 
the vector $\vec{v}^s$ is the velocity of the target which
is estimated from the history of the observations coming from $s$ but, depending on the sensor capabilities, it can be directly provided by $s$. 

We can now compute the \emph{topological mass} of each node in the topology, i.e., 
\begin{equation}
    M_n = \sum_{\forall i : \acute{q}^i_{t+1} = n} w^i = \sum_{\forall i : \acute{q}^i_{t+1} = n} w^i_q + w^i_{\vec{v}} .
\end{equation}
Then, the current best estimate of our target location is
\begin{equation}
    n_{t{+1}}^* = \underset{n \in N}{\text{argmax}} \; M_n
\end{equation}

\paragraph{Resampling}
We sample the new set of particles $P_{t{+1}}$, drawing from $\acute{P}_{t+1}$, with a probability distribution $f_w$ equal to the normalized weights
\begin{equation}
    p_{t{+1}}^i \sim f_w(i) = \frac{w^i}{\sum_{j=1}^{|\acute{P}_{t+1}|}{w^j}}. 
\end{equation}
Once the new set of particles has been sampled, we add noise to their states
\begin{equation}
    \begin{split}
        &q_{t+1} =  q_{t+1} \sim \mathcal{U}(N), \quad \text{with probability $\textrm{pr}_j$},\\
        &\vec{v}_{t+1} = \vec{v}_{t+1} + \epsilon_v, \quad \epsilon_v \sim \mathcal{N}(\mu_{noise},  \sigma^2_{noise}),\\
        &\tau_{t{+1}} = \tau_{t{+1}} + \epsilon_{\tau}, \quad \epsilon_{\tau} \sim \mathcal{U}(\tau^-_{noise}, \tau^+_{noise}). 
    \end{split}
\end{equation}

\section{The Next-Best-Sense Approach}\label{sec:nbs}
To further improve the estimate of the TPF, we adopted a decision-making framework for planning the next robot position to be the closest to the estimated picker location or to allow meaningful readings. With this in mind, we chose to use  Next-Best-Sense~\cite{polvara2020next}, an iterative exploration algorithm that executes sensing actions and updates a belief map with information from the most recent observation. Full details on the approach are reported in the original publication, we try now to summarize the most important bits of information.
NBS aims at planning the next robot pose by evaluating multiple \textit{criteria} into a single utility function. This is performed by using the MCDM method~\cite{mcdm} and the \textit{Choquet Fuzzy Integral} to model relations of redundancy and synergy existing among the criteria. For the application of the study, it has been decided to consider four criteria. 
    (i) \textit{Travel distance (TD)}, the distance between the current robot node and the candidate node. 
    (ii) \textit{Sensing time (ST)}, the time required for a sensing operation.
    (iii) \textit{RFID Information gain (RFID)}, the amount of entropy in the map after the robot reaches that node.
    (iv) \textit{Battery Status (BS)}, the expected battery level after completing the navigation task.
Each criterion is associated with a positive weight $\eta$ which defines its relative importance compared to the others. 
Being $K$ the set of all criteria, for a given pose $c$, we first sort the criteria according to their utility $u_i$ such that
    $u_{(1)}(c) \le ... \le u_{(|N|)}(c) \le 1$.
Furthermore, we define the set
$A_{(j)} = \{i \in N | u_{(j)}(c) \le u_{(i)}(c) \le u_{(|N|)}(c)\}$, 
the set of all criteria with utility larger than the utility from $k_j$. 
The global utility function $f(u_c)$ of a candidate pose \textit{c} is computed by a \textit{discrete} Choquet integral which uses the ranking of the utilities as well as the sets $A_j$, i.e.,

\begin{equation}\label{choquet} 
    f(c) = \displaystyle \sum_{j=1}^{|K|}(u_{(j)}(c) - u_{(j-1)}(c))\eta(A_{(j)}).
\end{equation} 

After all the nodes have been evaluated, NBS selects the next robot goal as the node with the highest utility function. Finding the best set of weights for the criteria is out of the scope of this paper. Based on the study shown in \cite{polvara2020next}, we identified a combination of weights working for the current application and we kept it fixed across all the experiments reported in the following section.

\section{Experiments}\label{experiments}

To validate the proposed architecture, we designed three sets of simulated experiments. Unfortunately, no in-field experimentation has been possible due to the COVID limitations in the UK. For this reason, we modeled the University of Lincoln's Riseholme campus in Gazebo, as shown in Fig. 2 (top). It consists of two 30 meter long polytunnels with five rows of raised beds and a storage and packaging facility nearby. Each row counts ten topological nodes and, overall, there are 137 nodes on the map. An overview of the topological map is given in Fig. 2 (bottom). 

\begin{figure}
  \centering
  {\includegraphics[width=\columnwidth]{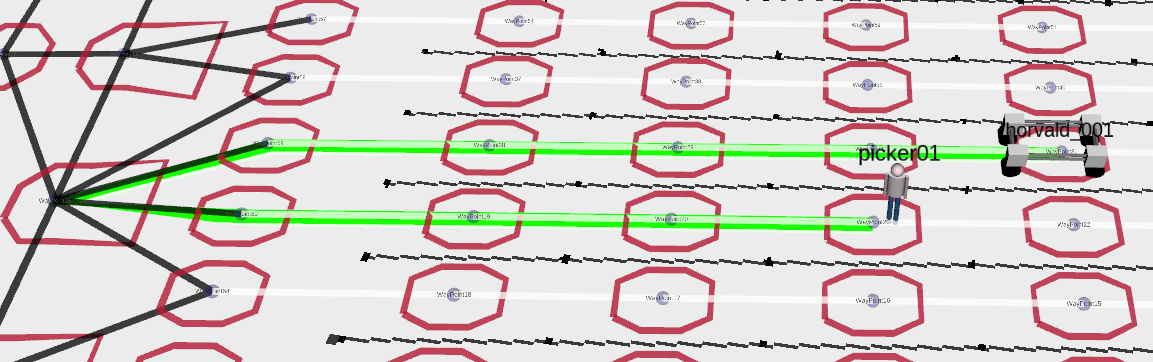}}
  \caption{A robot which, after reaching a picker in the wrong tunnel, has to re-plan a path going out of the current tunnel first to enter in the correct one.}
  \label{fig:replanning}
\end{figure}

In our simulation, we have one or more human agents traversing the polytunnels similarly to how they would do in the real setting: they enter the tunnels (outlined by black edges in Fig. 2 (bottom)) from the left side, follow the lane, and, upon arriving at the other end, they enter in an adjacent one proceeding in the opposite direction.
The humans traverse the same lane always in the same direction but, at any step, they have $p\_change$ chances of changing its direction of motion for $t\_reverse$ seconds (see Table \ref{tab:parameters}).
Each picker is equipped with a Waveshare SIM7600E-H GNSS unit for geolocalization. It is a low-cost device that has weaker performance than RTK-GNSS sensors. We empirically calculated the errors affecting the GPS signal 
which are then injected into our simulations. In particular, we identified the following type of noise:
    (i) a constant bias over the Euclidean coordinates of $gps\_offset$;
    (ii) a Gaussian white noise added to the signal at every step with variance $gps\_var$;
    (iii) a drifting over time from the ground truth position of $gps\_drift$; and
    (iv) a communication blackout of $gps\_off$ minutes (see Table \ref{tab:parameters}).
In addition to the GPS device, a uniquely identified SMARTRAC Frog 3D RFID tag
is assigned to each picker.
The robot used in this study is the Thorvald robot, shown in Figure~\ref{fig:thorvald}, equipped with two Hokuyo 2D laser scanners located at opposite platform's corners, and an RFID antenna. The laser scanners are used for people perception, employing the off-the-shelf \textit{leg\_detector} ROS package
, and for robot self-localization and obstacle avoidance. 
In the experiments performed, we considered two metrics for assessing the performance of our solution in localizing the fruit pickers: the \textit{topological error} and the \textit{Euclidean error}. The first represents the number of nodes in the shortest path in the graph (i.e., the topological map) between the estimated node and the closest node to the real position of the fruit pickers (highlighted in Fig. \ref{fig:replanning}). The Euclidean error is the Euclidean distance between the estimated node projected on a metric map and the ground truth position of the picker. It's important to notice that, given the topology of the map constraining the robot's movements only along edges connecting two nodes, a smaller Euclidean error does not always correspond to a small topological error. 

For all the experiments presented in the following sections, we used fixed parameters for the NBS algorithm and the TPF, which are reported in Table \ref{tab:parameters} for clarify.
We initialize a TPF model for each simulated human to track their position. Each TPF receives observations from all the sensors and, in the absence of them, performs predictions at a rate \texttt{prediction\_rate}$:= \frac{1}{4}$ Hz to update the picker tracking. 
The velocity vector of the GPS data $\vec{v}^{GPS}$ is estimated with a simple average over the last 10 poses received from the sensor. Each GPS and RFID detection from the sensor is sent to the corresponding TPF as an \emph{identifying observation} (see Section \ref{sec:problem-form}); LIDAR detections are instead sent to all the filters indistinctly given they cannot uniquely identify a person. 

\begin{table}[]
\centering
\caption{Parameters used in all the experiments performed.  They are divided in (top) NBS criteria's weights; (center) TPF parameters (see Section \ref{alg:tpf-algorithm} for full description); (bottom) environmental parameters affecting human motion and GPS signal.}
\label{tab:parameters}
\begin{tabularx}{\linewidth}{XX}
\hline\hline
\multicolumn{2}{c}{Next-Best-Sense}                                                                                                                            \\ \hline
\multicolumn{1}{X|}{$\eta(TD)$ = 0.3}          & \multicolumn{1}{X}{$\eta(ST)$ = 0.1}   \\
\multicolumn{1}{X|}{$\eta(RFID)$ = 0.35}       & \multicolumn{1}{X}{$\eta(BS)$ = 0.25}  \\ \hline
\multicolumn{2}{c}{Topological Particle Filter}                                                                                                                \\ \hline
\multicolumn{1}{X|}{$\epsilon_{H}$ = 0.6}      & \multicolumn{1}{X}{$\epsilon_{JSD}$ = 0.975} \\
\multicolumn{1}{X|}{$\mu_{init}$ = $0.0$} &
\multicolumn{1}{X}{$\mu_{noise}$ = $0.0$} \\
\multicolumn{1}{X|}{$\sigma^2_{init}$ = $5\mathrm{e}{-2}$} &
\multicolumn{1}{X}{$\sigma^2_{noise}$ = $5\mathrm{e}{-4}$} \\
\multicolumn{1}{X|}{$\tau^-_{init}$ = 0.0} &
\multicolumn{1}{X}{$\tau^+_{init}$ = 1.0} \\
\multicolumn{1}{X|}{$\tau^-_{noise}$ = -0.1} &
\multicolumn{1}{X}{$\tau^+_{noise}$ = 0.1} \\
\multicolumn{1}{X|}{$\gamma_n^{LIDAR}$ = 0.25} & \multicolumn{1}{X}{$\gamma_n^{RFID}$ = 1}     \\
\multicolumn{1}{X|}{$\gamma_n^{GPS}$ = 1}      & \multicolumn{1}{X}{$\gamma_{\vec{v}}^{LIDAR}$ = 0}     \\
\multicolumn{1}{X|}{$\gamma_{\vec{v}}^{RFID}$ = 0} & \multicolumn{1}{X}{$\gamma_{\vec{v}}^{GPS}$ = 1}     \\ \hline
\multicolumn{2}{c}{Environmental parameters}                                                                                                                \\ \hline
\multicolumn{1}{X|}{$p\_reverse$ = 0.1}         & \multicolumn{1}{X}{$t\_reverse$ = 60[s]} \\
\multicolumn{1}{l|}{$gps\_offset\_\{x/y\}  \sim \mathcal{U}(0.0, 3.5)$[m]} & \multicolumn{1}{X}{$gps\_var$ = 0.1[m]}     \\
\multicolumn{1}{l|}{$gps\_drift \sim \mathcal{N}(0.0, 2.5)$[m]}  & \multicolumn{1}{X}{$gps\_off \sim \mathcal{U}(0.5, 1)$[min]}     \\ \hline\hline
\end{tabularx}
\end{table}

\subsection{Single picker tracking}\label{subsec:exp1}
In the first batch of simulations, we compare our new formulation against \cite{khan2020incorporating} that relied only on a noisy GPS signal. For this reason, no robot is used in \cite{khan2020incorporating} because it doesn't add any significant information. Because our method integrates multi-modal data, we show results obtained by mounting a 2D LIDAR and an RFID antenna on a moving robot. 

\begin{figure}
    \centering
    \includegraphics[width=\columnwidth]{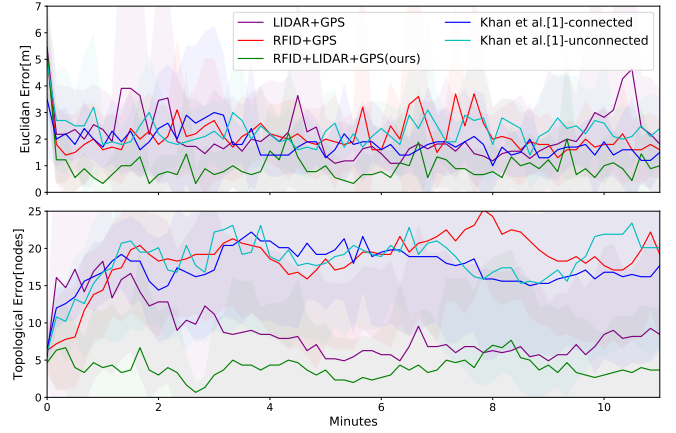}
    \caption{(Top) Euclidean Error (Top) and topological error (Bottom) - expressed as mean and standard deviation - while comparing our method and Khan \textit{et al.}~\cite{khan2020incorporating} in tracking a single fruit picker.}
    \label{fig:exp_1}
\end{figure}

We identified five different methods to compare: 
\begin{enumerate*}[label=(\roman*)]
    \item Khan \textit{et al.}~\cite{khan2020incorporating}-connected, in which the particle can only move between connected nodes;
    \item Khan \textit{et al.}~\cite{khan2020incorporating}-unconnected, in which the particles are allowed to move to nodes that are not directly connected with a small probability;
    \item LIDAR+GPS, in which a leg detector is combined with the GPS signal by the TPF;
    \item RFID+GPS, the TPF obtains observations from a GPS device and an RFID antenna;
    \item RFID+LIDAR+GPS(ours), in which all the three previously mentioned sensors provide observations to the TPF. 
\end{enumerate*}
\indent In order to provide more reliable results, for each method we performed ten runs and we averaged the results. 
These, expressed as mean and standard deviation, are reported in Figure~\ref{fig:exp_1} showing, despite all the methods are affected by a comparable Euclidean error, our solution combining RFID, LIDAR, and GPS signal outperforms all the others when considering the topological error. For the task at hand, this is a very significant result because a miss-prediction of where the picker is located would make the robot leave the row where it is currently located and traversing a long distance before reaching the human. Quantitative results are reported in Table~\ref{tab:results}, showing Euclidean and topological errors as mean absolute error along with the entire recorded task. Our method in the formulation using all the sensors proves to be more accurate than the compared ones, with an error of only $0.95(0.5)$ meters on a metric map and $3.89(1.42)$ nodes on the topological one. The latter result represents an improvement of $4.7\times$ compared to \cite{khan2020incorporating}, which has a topological error of $18.42(2.81)$ nodes. This can be interpreted such that, on average, there are eighteen nodes between the real position of the pickers and the robot, making the robot leave the row in which it is currently located in order to re-enter in the correct one.  We identify the ability of our TPF to adapt to variable target's speed as the main reason for its success against \cite{khan2020incorporating} in which the authors identified empirically a fixed $\lambda = 0.1$ (see Eq.~\ref{eq:prob_jump}), assuming all the tracked agents move at a constant speed. In addition to this, the ability of our method to fuse together multi-modal data coming from different sensors allows overcoming any noise and bias affecting the GPS signal and limiting \cite{khan2020incorporating}'s localization accuracy.

\begin{table}
\centering
\caption{Performance for each methods in all the experiments performed. Results are reported as mean and standard deviation. The best method is in bold.}
\label{tab:results}
\begin{tabularx}{\linewidth}{XXX}
\hline\hline
\multicolumn{1}{c|}{Method}           & \multicolumn{1}{c|}{\begin{tabular}{@{}c@{}}Euclidean \\ Error[m]\end{tabular}} &  \multicolumn{1}{c}{\begin{tabular}{@{}c@{}}Topological \\ Error[nodes]\end{tabular}}   \\ \hline\hline
\multicolumn{3}{c}{Single picker}                                                \\ \cline{1-3}
\multicolumn{1}{l|}{Khan \textit{et al.}[1]-unconnected} & \multicolumn{1}{c|}{2.21(0.44)}         &    \multicolumn{1}{c}{18.42(2.81)}                  \\
\multicolumn{1}{l|}{Khan \textit{et al.}[1]-connected}   & \multicolumn{1}{c|}{1.82(0.45)}         &    \multicolumn{1}{c}{17.59(2.46)}                  \\
\multicolumn{1}{X|}{LIDAR+GPS}       & \multicolumn{1}{c|}{2.08(0.82)}         &    \multicolumn{1}{c}{8.89(3.5)}                 \\
\multicolumn{1}{X|}{RFID+GPS}        & \multicolumn{1}{c|}{2.02(0.5)}        &    \multicolumn{1}{c}{18.86(3.57)}                 \\
\multicolumn{1}{X|}{RFID+LIDAR+GPS(\textit{ours})}  & \multicolumn{1}{c|}{\textbf{0.95(0.5)}} &    \multicolumn{1}{c}{\textbf{3.89(1.42)}}                  \\ \cline{1-3}
\multicolumn{3}{c}{Navigation Strategy}                                          \\ \cline{1-3}
\multicolumn{1}{X|}{EstimatedNode}      & \multicolumn{1}{c|}{2.21(1.96)}                &    \multicolumn{1}{c}{12.03(14.02)}                  \\
\multicolumn{1}{X|}{Next-Best-Sense\cite{polvara2020next}} & \multicolumn{1}{c|}{\textbf{0.95(0.5)}}                &    \multicolumn{1}{c}{\textbf{3.89(1.42)}}                  \\ \cline{1-3}
\multicolumn{3}{c}{Multiple pickers}                                              \\ \cline{1-3}
\multicolumn{1}{X|}{Dondrup et al.~\cite{dondrup2015real}}        & \multicolumn{1}{c|}{\textbf{1.66(1.2)}}                &    \multicolumn{1}{c}{17.37(13.89)}                  \\
\multicolumn{1}{X|}{RFID+LIDAR+GPS(\textit{ours})}   & \multicolumn{1}{c|}{1.94(2.34)}        &    \multicolumn{1}{c}{\textbf{9.82(12.57)}}                  \\
\multicolumn{1}{X|}{NoMonitor}   & \multicolumn{1}{c|}{3.48(4.23)}        &    \multicolumn{1}{c}{15.10(17.44)}                  \\
\multicolumn{1}{X|}{CostantSpeed}   & \multicolumn{1}{c|}{2.43(1.98)}        &    \multicolumn{1}{c}{13.62(14.94)}                  \\
\hline\hline
\end{tabularx}
\end{table}

\subsection{An intelligent robotic navigation framework}\label{subsec:exp2}
\begin{figure}
    \centering
    \includegraphics[width=\columnwidth]{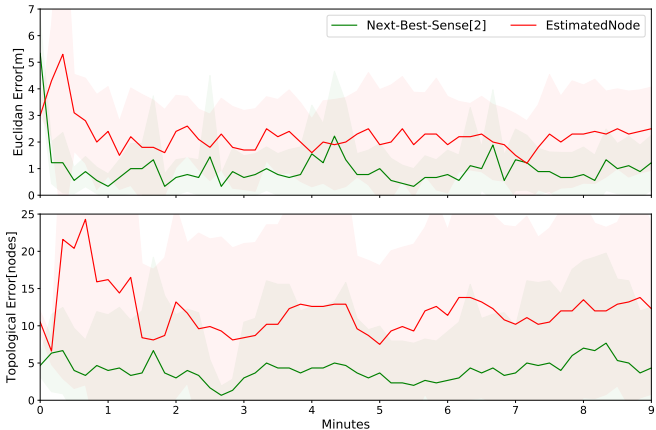}
    \caption{Comparison between Next-Best-Sense~\cite{polvara2020next} and EstimatedNode navigation strategies in localizing a single picker. 
    }
    \label{fig:exp_2}
\end{figure}

The second experiment we present shows the benefit of using an intelligent decision-making framework such as Next-Best-Sense to improve the fruit-pickers localization performance. Therefore, we compare NBS against a simpler policy, which is the one currently adopted in the real setting. More specifically, when a fruit picker calls a robot, the latter navigates towards the node estimated by the particle filter despite the filter's prediction accuracy. We call this policy \textit{EstimatedNode}. We recall that NBS combines instead multiple criteria in a single utility function and selects the next robot position greedily. The experimental setting is the same already discussed in subsection~\ref{subsec:exp1}, with a single picker traversing the polytunnels. 
The performance is shown in Figure~\ref{fig:exp_2} and is reported in Table~\ref{tab:results}, in which it is possible to see NBS outperforms EstimatedNode for both the metrics considered. In particular, NBS is $2.32\times$ and $3.09\times$ more accurate than EstimatedNode concerning the Euclidean error and the topological error, respectively. 
We recall that, in this scenario, accuracy is associated with a smaller probability to estimate the wrong lane in the polytunnel, leading the robot to re-plan a path with a noticeable expense of time and resources, similarly to what can be seen in Figure~\ref{fig:replanning}.

\subsection{Multi-pickers tracking}\label{subsec:exp3}
\begin{figure}
    \centering
    \includegraphics[width=\columnwidth]{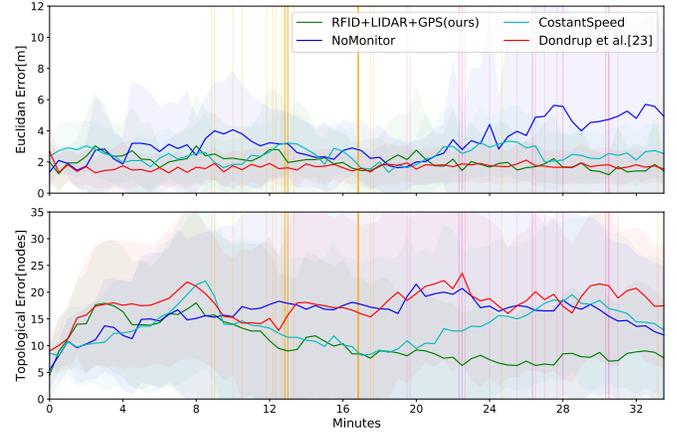}
    \caption{Comparison between the proposed method with Dondrup \textit{et al.}~\cite{dondrup2015real} in localizing and tracking three pickers. 
    Vertical bars indicate that a picker is a the entrance (pink) or end (orange) of a tunnel and it is switching lane (and direction).}
    \label{fig:exp_3}
\end{figure}
In the last batch of experiments, we present a more complete scenario in which a single robot serves multiple fruit pickers. 
We place 3 fruit-pickers in our simulated world, each of them characterized by his motion and independent noise, as described in Sec. \ref{experiments}.
Here, differently from the single picker scenario, false-positive detections from the LIDAR can easily mistake one person for another, especially if they are in a nearby tunnel and/or the GPS error is very large. Moreover, the short-range sensors on the robot cannot detect all the pickers around the field at each given moment; therefore, the ability to predict from sparse observations is here more important than in the single-picker case.

In this experiment, we offer a comparison with a state-of-the-art human tracking system, the \textit{bayesTracker} library presented in \cite{dondrup2015real}. Similar to our approach, \textit{bayesTracker} is a mobile robotics framework able to track the position of multiple people moving in the environment and it is based on multiple sensor observations. This method works at the metric level and does not exploit the topological structure of the environment, as our proposed approach does. It combines the GPS signal and the LIDAR's based leg detector via an Unscented Kalman Filter and a constant velocity model to estimate and track the people's location. 
However, its use is limited to pose-like input data and does not accept the likelihood distributions provided by the RFID sensor, unlike our TPF. 
The \textit{bayesTracker} outputs a list of tracked people identified by UUIDs to maintain the tracking consistency over time. However, at times it can return more than one tracking UUID for each person -- e.g., when it is not able to match the detections from the different sensors -- or less than one
-- e.g., when the GPS signal is not available for a certain time. 
To compare with the results of our TPF, the output from \textit{bayesTracker} is parsed to obtain at most one tracking for each picker at each given time. Each GPS ID, which uniquely identifies the pickers, is initially associated with the closest (in Euclidian distance) \emph{bayesTracker}'s UUID, and the association is maintained until the tracked UUID is lost. At this point, it gets associated with the closest of the UUIDs not already assigned to other pickers.

To study the contributions of each improvement to the TPF introduced here, 
compared to \cite{khan2020incorporating}, we show an ablation study of our proposed method.
In particular, the method named \textit{CostantSpeed} does not estimate the particles velocity but uses a constant value, similar to \cite{khan2020incorporating}. The method \textit{NoMonitor}, instead, is stripped of the dual mechanism monitoring the divergence from observations and the entropy of the distribution of the particles (see Section \ref{sec:tpf}).

We plotted the average errors among $3$ pickers for $10$ runs in Fig. \ref{fig:exp_3}, while numerical data are reported in Table \ref{tab:results}. 
Not surprisingly, \cite{dondrup2015real} leads to the smallest Euclidean error, while, at the same time, the largest topological error. This is somewhat justified by the fact that this method largely depends on the noisy GPS signal.
The \emph{bayesTracker} predictions can be relatively close to the ground truth on the metric map but in the wrong lane of the polytunnel, resulting in a very large topological error. Our proposed method achieves the best overall performance, reporting metric errors very close to \cite{dondrup2015real} (less than $0.3m$) despite being designed to work on topological maps. 
Both the ablated methods perform worse than the proposed approach.
From Fig. \ref{fig:exp_3} we can observe how \emph{NoMonitor}'s Euclidean error increases significantly in the second half of the plot, where the pickers are returning near the entrance of the tunnels. Here, without our monitoring mechanism, the particles spread more easily with the increased connectivity between nodes. Moreover, the topological error of \emph{ConstantSpeed} increases in the second half of the plot when the pickers are changing their direction of movement, whereas the complete method can maintain a low error. The observations suggest that the method's performance can be susceptible to the structure of the topology, particularly in terms of the connectivity between nodes.

\section{Discussion and Conclusion} \label{conclusion}
In this paper, we present a robotic framework for localizing and tracking people in a structured environment. It combines an improved formulation of the topological particle filter (TPF) with a multi-criteria optimization path planner such as Next-Best-Sense (NBS)~\cite{polvara2020next}. NBS is used to identify new robot positions which can lead to obtaining meaningful observations for improving the TPF estimate on the people's location. 
Compared to the previous work~\cite{khan2020incorporating}, our revised TPF allows to integrate multi-modal sensors readings, and it estimates the velocity of the tracked people allowing us to localize targets moving at a variable speed under noisy observations. We adopted a mechanism to monitor the distribution of the particles with the goal of preventing it from diverging from the true target position.   
The improvements we introduced lead to an increase in localization accuracy equal to $4.7\times$ compared to \cite{khan2020incorporating}. We also show how planning intelligent robot sensing poses over the topological map can further improve the results compared to a more straightforward navigation policy such as moving towards the estimated node. 
Although our approach performance is limited by the lower bound set by the average length of the edges connecting the nodes in the topology, we show that in a polytunnel-like structured environment it is more effective than approaches based on continuous space tracking~\cite{dondrup2015real} at estimating the location of humans. In future work, we plan to assess how the performance is affected by other topologies -- rather than only polytunnels -- and to validate the approach in the real-world setting in the polytunnels at the University of Lincoln.
Moreover, we aim to integrate the targets' identities in the particles state to allow tracking multiple people more efficiently. 

\bibliography{biblio}
\bibliographystyle{ieeetr}

\end{document}